\documentclass{article}
\usepackage{spconf,amsmath,graphicx}

\usepackage{enumitem}
\setlist{nosep, leftmargin=14pt}

\usepackage{mwe} % to get dummy images
\usepackage{multirow}
\usepackage{booktabs}
\usepackage{subcaption}
\title{\textit{CTest-Metric:} A Unified Framework to Assess Clinical Validity of Metrics for CT Report Generation}
\name{Vanshali Sharma, Andrea Mia Bejar, Gorkem Durak, and Ulas Bagci\thanks{This research is partially supported by the NIH grant R01-HL171376.}}
\address{Machine and Hybrid Intelligence Lab, Northwestern University, Chicago, IL, USA}
\begin{document}
%\ninept
%
\maketitle
In the generative AI era, where even critical medical tasks are increasingly automated, radiology report generation (RRG)  continues to rely on suboptimal metrics for quality assessment. Developing domain-specific metrics has therefore been an active area of research, yet it remains challenging due to the lack of a unified, well-defined framework to assess their robustness and applicability in clinical contexts. To address this, we present \textbf{\textit{CTest-Metric}}, a first unified metric assessment framework with three modules determining the clinical feasibility of metrics for CT RRG. The modules test: (i) Writing Style Generalizability (WSG) via LLM‑based rephrasing; (ii) Synthetic Error Injection (SEI) at graded severities; and (iii) Metrics‑vs‑Expert correlation (MvE) using clinician ratings on 175 “disagreement” cases. Eight widely used metrics (BLEU, ROUGE, METEOR, BERTScore‑F1, F1‑RadGraph, RaTEScore, GREEN Score, CRG) are studied across seven LLMs built on a CT‑CLIP encoder. Using our novel framework, we found that lexical NLG metrics are highly sensitive to stylistic variations; GREEN Score aligns best with expert judgments (Spearman~0.70), while CRG shows negative correlation; and BERTScore‑F1 is least sensitive to factual error injection. 
We will release the framework, code, and allowable portion of the anonymized evaluation data (rephrased/error-injected CT reports), to facilitate reproducible benchmarking and future metric development.
 %Using this novel framework, we showed that widely adopted standard evaluation metrics in existing studies fail to capture the finer intricacies of CT reporting and demonstrated significantly low Spearman's rank correlation (ranging from -0.27 to 0.7) with Expert Rating.

%Despite being an active area of research, developing domain-specific metrics remains challenging due to the lack of a unified, well-defined framework to assess their robustness and applicability in clinical contexts. Therefore, we present \textbf{\textit{CTest-Metric}}, a first unified metric assessment framework  that can be utilized to determine the clinical feasibility of metrics for CT report generation. In this context, our framework reveals that existing CT report generation studies predominantly rely on metrics developed for X-ray report evaluation, failing to capture the finer intricacies of CT reporting and showing significantly low Spearman's rank correlation (ranging from -0.27 to 0.7) with Expert Rating. We will release the framework, code and partially anonymized evaluation data (rephrased/error-injected CT reports), to facilitate reproducible benchmarking and future metric development.

\section{Introduction}

\begin{figure*}[h]
\centering     
     \includegraphics[width=0.9\linewidth]{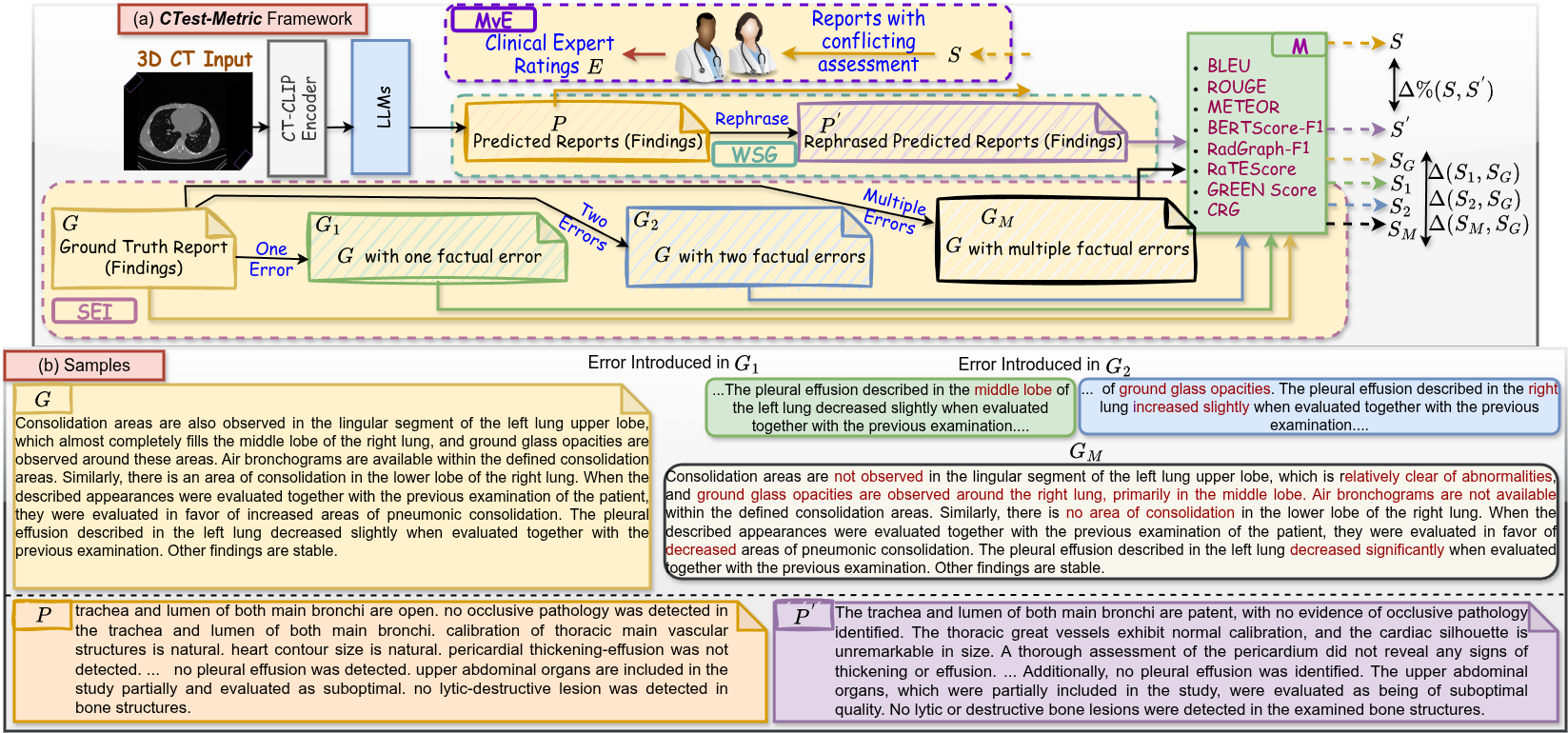}
    \caption{(a) The propsoed framework, \textit{CTest-Metric}, comprises three modules: (i) Writing Style Generalizability Test (WSG); (ii) Synthetic Error Injection Test (SEI); and (iii) Metrics‑vs‑Expert Correlation Test (MvE). (b) Sample reports are given.}
    \label{fig:framework}
\end{figure*}
%to evaluate whether automated metrics used for CT radiology report generation (RRG) are clinically meaningful

With recent breakthroughs in Large Language Models (LLMs) for automated radiology report generation (RRG)~\cite{xin2025med3dvlm, di2025ct}, a crucial question arises: \textit{Do existing metrics capture what actually matters to trace the clinical efficacy and acceptability of LLM-generated reports?} A clinically grounded metric must be insensitive to report writing styles and sensitive to subtle yet critical factual mismatches. It should also identify synonymous medical terminologies, and be robust enough to capture overfitting in AI-based generative models before deploying them for clinical use.

Despite the availability of a large number of metrics, most are developed for general-domain text~\cite{papineni2002bleu, lin2004rouge, banerjee2005meteor}, and even a few domain-specific metrics overlook deeper clinical relevance or ignore synonymous terminologies. Therefore, the existing RRG methods~\cite{di2025ct, shi2024med} continue to rely on metrics that are not reflective of clinical fidelity. Consequently, the designing of radiology-specific metrics has been an active area of research with the aim of evaluating both clinical aspects and linguistic similarity between generated radiology reports and the ground truth. However, due to the absence of any standardized tool and well-defined criteria to test these metrics, RRG tasks still rely on inconsistent metrics, resulting in misleading model selections. This underscores the need for a system that can serve as a well-defined framework for metrics developers to assess the clinical applicability of the metrics.

Few prior studies~\cite{banerjee2024rexamine,yu2023evaluating} have addressed this problem; however, their focus has predominantly been on the X-ray RRG. Since X-ray images are 2D scans, the corresponding reports are limited to a specific anatomical context and span shorter sentences. In contrast, 3D CT scans capture volumetric multi-slice information, resulting in semantically denser narratives and a broader vocabulary that includes richer anatomical detail, diverse medical terminology, lesion descriptions, and measurements. Most existing clinical-efficacy (CE) metrics were originally tailored to X-ray-based vocabulary and thus struggle to capture complex and diverse terminologies present in CT reports. Despite this limitation, CT-based studies still use these metrics for reporting results and model comparison.  Therefore, CT RRG requires special attention in terms of both designing appropriate metrics and developing a tool to assess their applicability and feasibility.

In this paper, we develop \textit{\textbf{CTest-Metric}}, a framework for evaluating the extent to which a given metric satisfies the criteria for being clinically grounded. This assessment is conducted on eight benchmarking metrics that have been extensively used in recent CT report generation studies, ensuring consistency with established evaluation practices. The proposed framework includes three analytical modules: a) \textit{\textbf{Writing Style Generalizability Test (WSG)}} examines the metrics' generalizability across different writing styles, b) \textbf{\textit{Synthetic Error Injection Test (SEI)}} introduces factual errors in the reports at three different levels and investigates their impact on the metrics' outcomes, and c) \textit{\textbf{Metrics-vs-Expert Correlation Test (MvE)}} obtains expert ratings for reports exhibiting disagreement among the metrics. Further, a correlation is established between the eight metrics and expert ratings for a comprehensive study. By leveraging reports generated using seven different LLMs in conjunction with expert assessment, the proposed framework presents a robust pathway that metrics developers can utilize when designing new metrics.      
The paper's contributions are summarized below:
\begin{itemize}
    \item \textbf{First framework to assess metrics for CT RRG}: We developed \textit{\textbf{CTest-Metric}}, a novel unified framework to assess metrics for CT RRG. It investigates eight benchmarking metrics on CT reports generated by seven different LLMs and analyzes the behavior of both NLG (text-based) and CE metrics for CT report evaluation.
    %(commonly used for CT but originally intended for X-ray reports)
    \item \textbf{Expert assessment}: Our study selected 175 cases across predictions from seven LLMs where the metrics showed conflicting assessments. These specific reports were then reviewed by clinical experts, and correlations were derived between expert ratings and the metrics' scores.
    \item \textbf{Analyzed the impact of stylistic variations and graded synthetic errors}: We introduced stylistic variations and factual errors at three severity levels in the CT reports. %These modifications were performed using an LLM guided by zero-shot instruction-based prompting. 
    We quantified the effect of these changes on metric sensitivity. 
\end{itemize}

\section{Related Work}
%\subsection{Metrics Development and  for Radiology Report Generation}
The RRG literature reveals that various NLG and CE metrics are commonly used to assess predicted CT reports. The earlier studies primarily relied on NLG metrics, including BLEU~\cite{papineni2002bleu}, ROUGE~\cite{lin2004rouge}, METEOR~\cite{banerjee2005meteor}, and BERTScore-F1~\cite{zhang2019bertscore}. These metrics quantify textual similarity, for example, BLEU measures n-gram overlap, ROUGE emphasizes sequence-level recall, and METEOR incorporates synonym matches between the generated and the ground truth report. Similarly, BERTScore-F1 computes similarity using contextual embeddings. To validate the clinical context, CE metrics including  F1-RadGraph~\cite{jain2021radgraph} and CheXpert~\cite{irvin2019chexpert} were introduced. While the former extracts entities and relations from the given reports and measures graph-level scores, the latter adopts a rule-based labeler for 14 chest X-ray findings. More recently, RaTEScore~\cite{zhao2024ratescore}, GREEN Score~\cite{ostmeier2024green}, and CRG~\cite{hamamci2025crg} were introduced. The RaTEScore compares reports on the entity embedding level and the GREEN score uses regular expressions to parse error counts from their pre-trained model output. Unlike other score, the CRG balances penalities based on label distribution in the reports and ignores clinically irrelevant true negatives.
% resulting in an increasing number of studies~\cite{liu2025enhanced, nath2025vila} incorporating these metrics.

For the X-ray RRG tasks, prior works~\cite{wang2023r2gengpt, chen2025dia} followed a similar trend, relying on the same set of standard metrics including the four BLEU n-gram levels, METEOR, ROUGE, and CheXpert. A largely identical evaluation strategy is also observed in CT RRG, where most studies~\cite{xin2025med3dvlm, di2025ct} adopt a similar set of NLG and CE metrics, despite the latter being X-ray-oriented and thus, do not fully capture the semantic and anatomical complexity of CT reporting. Although the literature lacks any unified framework for evaluating CT RRG metrics, some X-ray-focused works such as Yu et al.\cite{yu2023evaluating} and Banerjee et al.~\cite{banerjee2024rexamine}  highlighted that even advanced metrics can be inconsistent and poorly correlated with expert ratings.

%remain the most widely used, some papers report RaTEScore~\cite{zhao2024ratescore} under CE metrics. 
%\subsection{Metrics Development and Assessment Platforms }

\section{Methodology}
\textbf{Overview.} Given a 3D CT scan $i\in I$ with corresponding ground truth report $r\in R$, we employed seven report generation models to generate CT report $p^j_i \in P$, where $j={1,2,..,7}$. The seven deep learning models adopt a CT-CLIP~\cite{hamamci2024foundation} image encoder to extract image features combined with seven different LLMs, including variants of GPT (Distilgpt, GPT2, GPT2-Medium, LLaMA-3.2-1B) and LLaMA (LLaMA-3.2-1B, LLaMA-2-7b-chat-hf), with a biomedical-domain LLM (BioGPT-Large). These variants are paired with two configurations: LLM fine-tuning\footnote{\label{clip-text-decoder}https://github.com/fkodom/clip-text-decoder} and a frozen-LLM setup (R2GenGPT's shallow alignment~\cite{wang2023r2gengpt}).
%including , R2GenGPT~\cite{wang2023r2gengpt}, R2GenGPT with BioGPT-Large, and R2GenGPT with . 
%The seven selected LLMs 

The predictions obtained from each model are assessed using a set of eight metrics $M$, where $M$= \{BLEU, ROUGE, METEOR, BERTScore-F1, F1-RadGraph, RaTEScore, GREEN Score, CRG\}. The first four metrics are NLG-based whereas the last four are designed for clinical-efficacy check. The proposed \textbf{\textit{CTest-Metric}} framework is illustrated in Fig.~\ref{fig:framework}, comprising three analytical modules, as detailed below.

%Note that these metrics were chosen because they are widely adopted in recent CT report generation literature, ensuring consistency with established evaluation practices.

\subsection{Writing Style Generalizability Test (WSG)}
\label{wsg}
In the WSG module, metrics are tested for sensitivity to report writing style. It evaluates whether they exhibit significant performance shifts in response to variations in stylistic modifications despite unchanged clinical semantics. This test employs an LLM-based approach using zero-shot instruction-based prompting on LLaMA-3.1-8B-Instruct. This approach rephrases the predicted reports $P$ generated by the seven models into $P'$, while preserving clinical outcomes and semantics. Each metric is applied on $P$ and $P'$, and the final difference is analyzed. A lower difference indicates metrics' robustness towards stylistic variations. 

\subsection{Synthetic Error Injection Test (SEI)}
The SEI module monitors how the metrics penalizes factual errors of varying severity. We study three levels of errors intentionally injected using the same prompting technique used in Sec \ref{wsg}. It involved introducing one, two and multiple synthetic errors in ground truth reports, followed by evaluation at each level using the eight metrics from $M$. Specifically, it examines the deviation of scores ($\Delta(S_j, S_G)= S_j - S_G$) between the ideal case $S_G$ (where the ground truth is evaluated against itself) and report scores $S_j$ with intentionally injected errors at level $j$.  A substantial difference in metrics' scores for varying levels of discrepancy shows its strong discriminative ability to capture factual inconsistencies.

\subsection{Metrics-vs-Expert Correlation Test (MvE)}
\label{sec:mve}
In this module, we systematically quantified the correlation between metrics in $M$ and expert ratings $E$. We also examined inter-metric correlations to understand how they agree or disagree in report quality evaluation. We computed this correlation on 175 patient reports for which automated metrics provided conflicting assessments. To this end, we initially derived per-patient scores for all eight metrics corresponding to each LLM. After normalizing the per-patient scores, we calculated the standard deviation across all metrics to assess disagreement. The top 25 cases with the highest standard deviation (i.e, highest variable) were selected. These 175 cases were finally reviewed by clinical experts, yielding expert ratings $E$. We then quantified the alignment by computing Spearman’s rank correlation coefficient ($\rho$) between every pair of metrics (for inter-metric correlation) and between each metric in $M$ and the expert ratings $E$. A higher $\rho$ signifies that the two metrics tend to rank patients in a similar order while a lower $\rho$ shows disagreement in their assessment.

%A high standard deviation implies a higher degree of disagreement among metrics.

\section{Experiments and Results}
\subsection{Dataset and Training Details}
We used a publicly available 3D medical imaging dataset, CT-RATE~\cite{hamamci2024foundation}. It comprises 50,188 non-contrast chest CT volumes, along with their corresponding radiology reports. We used their official training and validation split. All RRG models were trained for 10 epochs on NVIDIA A100 GPU using the hyperparameters specified in the publicly released code.

\begin{figure*}[t]
    \centering
    
    \begin{subfigure}[t]{0.49\textwidth}
        \centering
        \includegraphics[width=\linewidth]{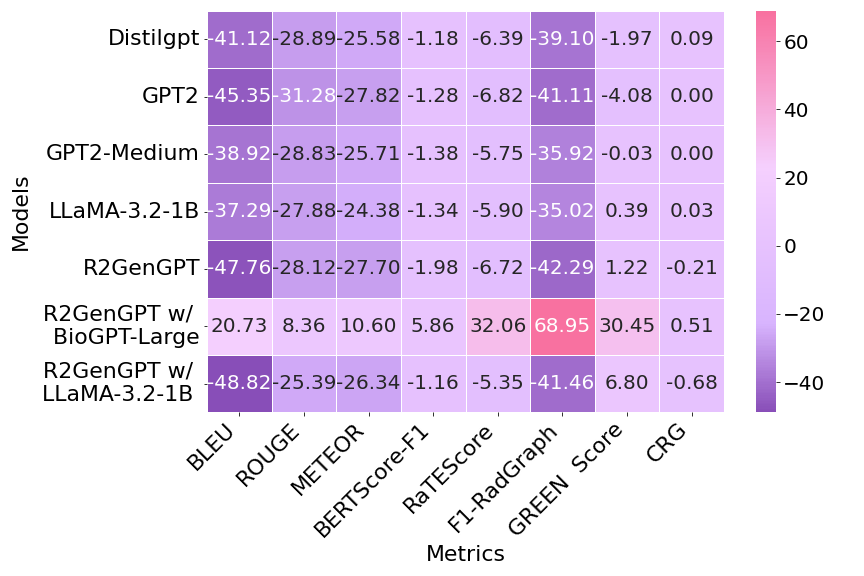}
        \caption{Rephrasing robustness (WSG).}
        \label{fig:rephrase}
    \end{subfigure}
    \hfill
    \begin{subfigure}[t]{0.49\textwidth}
        \centering
        \includegraphics[width=\linewidth]{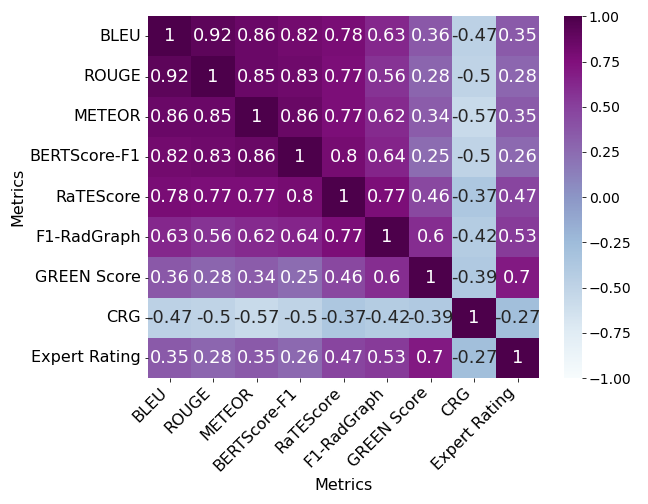}
        \caption{Inter-metric and expert correlation.}
        \label{fig:spearman}
    \end{subfigure}

    \caption{Evaluation of metric reliability and robustness across rephrasing, factual error severity, and metric-expert correlations.}
    \label{fig:three_panel_results}
\end{figure*}

\subsection{Analyzing Writing Style Generalizability Test (WSG)}
The WSG results are presented in Fig.~\ref{fig:rephrase}, where each heatmap cell value represents the percentage difference ($\Delta\%$) between the original score ($S$) and the score obtained after rephrasing ($S^{'}$) for the corresponding model-metric pair. It can be observed that NLG-based metrics were the most impacted by the stylistic variations (ranging from -48.82\% to -1.16\%) because they primarily measure lexical overlap rather than underlying clinical semantics. Among CE metrics, F1-RadGraph experienced the most significant deviation (ranging from -42.29\% to 68.95\%) indicating its high sensitivity to rephrasing. The best two performing metrics include the CRG and the GREEN Score. Since all CE metrics except CRG were introduced focusing on X-Ray datasets, their scores are likely to shift substantially on a vast terminology set of CT reports, especially when rephrased. Although built on X-Ray corpora, the GREEN score was originally tested on out-of-domain modalities, including CT scans, making it relatively robust. Consequently, it performs comparably to CRG, which was developed for CT reports. CRG considers the presence/absence of multi-label clinical entities instead of sentence-level structure, making it less susceptible to performance shift under rephrasing.  

\subsection{Analyzing Synthetic Error Injection Test (SEI)}
\begin{figure}[h!]
    \centering
    \includegraphics[width=\linewidth]{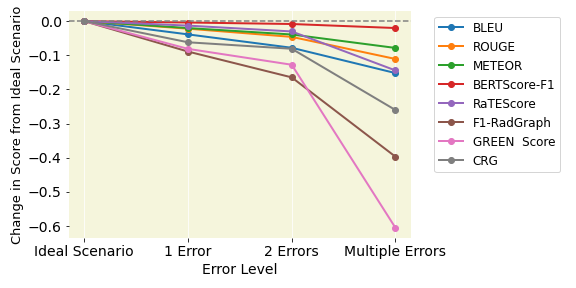}
    \caption{Metric response across increasing error levels.}
    \label{fig:errorlevel}
\end{figure}

The SEI results in Fig.~\ref{fig:errorlevel} signify that the metric least impacted by synthetic errors is the BERTScore-F1 with $\Delta (S_M, S_G)$ of -0.02 (approximately -2\%), whereas the most affected metrics include the GREEN Score ($\Delta(S_1, S_G)$ = -0.0812, $\Delta(S_2,S_G)$ = -0.1277, $\Delta(S_M,S_G)$ = -0.6053) and the F1-RadGraph ($\Delta(S_1, S_G)$ = -0.0891 , $\Delta(S_2,S_G)$ = -0.1647, $\Delta(S_M,S_G)$ = -0.3970). It can be observed that all metrics exhibit similar behavior when the report has minimal difference from the ideal case but the performance significantly diverges when multiple errors are injected. Therefore, it is crucial to examine various error levels to identify metrics' sensitivity to critical factual errors.  

\subsection{Analyzing Metrics-vs-Expert Correlation Test (MvE)}
%The original scores $S$ on the test set across all automated metrics are reported in Table \ref{tab:original_score}. However, 
As discussed in Sec. \ref{sec:mve}, we considered 175 conflicting cases for this test. As shown in Fig.~\ref{fig:spearman}, the GREEN Score closely resembles the Expert Rating $E$ with $\rho$ = 0.7 followed by F1-RadGraph with $\rho$ = 0.53. The NLG-based metrics demonstrated a similar correlation trend with $E$, reporting $\rho$ in a narrow range of 0.26 to 0.35. The worst performance was observed using the CRG, which presented a negative correlation with $E$ and other metrics. A high correlation range among NLP-based metrics indicates strong agreement in their assessment. While F1-RadGraph exhibits consistent alignment to other metrics, it shows a slight bias to NLG-based metric outcomes rather than the GREEN Score and Expert Rating. 

\section{Discussion and Concluding Remarks}
This study introduces \textit{CTest‑Metric}, a modular framework to evaluate automated metrics used in CT RRG, assessing their ability to capture clinical context in reports. By considering style robustness (WSG), factual‑error sensitivity (SEI), and alignment with expert judgment (MvE), the work demonstrates that widely used NLG metrics are brittle to rephrasing and incompletely capture factual correctness, whereas CE metrics vary substantially in their agreement with experts. In particular, GREEN Score exhibits the highest association with expert ratings on a curated set of ``disagreement" cases, while CRG remains robust to stylistic changes but correlates negatively with expert scores. This can be attributed to its reliance on label-level information which makes it largely insensitive to rephrasing, yet it remains less aligned with expert evaluations. These findings have immediate implications for metric choice in CT RRG and for how future metrics should be stress‑tested before deployment. 

%The MvE study design is diagnostic but may bias absolute correlation estimates relative to a random test sample. 
Expert validation involved two reviewers, with a second opinion sought for ambiguous cases, which limits the diversity of assessment. WSG and SEI depend on an LLM‑based prompting, which primarily introduced laterality and negation errors, though the fidelity of these edits was not independently validated.    
%Details of the expert rating protocol (panel size, rubric, inter‑rater agreement) are not reported here, constraining assessment of annotation reliability. WSG and SEI depend on an LLM‑based paraphraser/editor; while prompts instruct semantic preservation and controlled error injection, the fidelity of edits was not independently validated. 
Finally, all experiments are conducted on CT‑RATE (one of the largest in the literature); broader generalization to other institutions, contrast phases, or body regions remains to be established. These constraints highlight opportunities for expanded validation and reporting in subsequent versions.

\section{Compliance with ethical standards}
\label{sec:ethics}
This research study was conducted retrospectively using human subject data made available in open access by (CT-Rate). Ethical approval was not required as confirmed by the license attached to the open-access data.

\bibliographystyle{IEEEbib}
\bibliography{strings,refs}

@article{xin2025med3dvlm,
  title={Med3dvlm: An efficient vision-language model for 3d medical image analysis},
  author={Xin, Yu and Ates, Gorkem Can and Gong, Kuang and Shao, Wei},
  journal={arXiv preprint arXiv:2503.20047},
  year={2025}
}

@inproceedings{di2025ct,
  title={Ct-agrg: Automated abnormality-guided report generation from 3d chest ct volumes},
  author={Di Piazza, Theo and Lazarus, Carole and Nempont, Olivier and Boussel, Loic},
  booktitle={2025 IEEE 22nd International Symposium on Biomedical Imaging (ISBI)},
  pages={01--05},
  year={2025},
  organization={IEEE}
}

@inproceedings{papineni2002bleu,
  title={Bleu: a method for automatic evaluation of machine translation},
  author={Papineni, Kishore and Roukos, Salim and Ward, Todd and Zhu, Wei-Jing},
  booktitle={Proceedings of the 40th annual meeting of the Association for Computational Linguistics},
  pages={311--318},
  year={2002}
}

@inproceedings{lin2004rouge,
  title={Rouge: A package for automatic evaluation of summaries},
  author={Lin, Chin-Yew},
  booktitle={Text summarization branches out},
  pages={74--81},
  year={2004}
}

@inproceedings{banerjee2005meteor,
  title={METEOR: An automatic metric for MT evaluation with improved correlation with human judgments},
  author={Banerjee, Satanjeev and Lavie, Alon},
  booktitle={Proceedings of the acl workshop on intrinsic and extrinsic evaluation measures for machine translation and/or summarization},
  pages={65--72},
  year={2005}
}

@article{shi2024med,
  title={Med-2e3: A 2d-enhanced 3d medical multimodal large language model},
  author={Shi, Yiming and Zhu, Xun and Wang, Kaiwen and Hu, Ying and Guo, Chenyi and Li, Miao and Wu, Ji},
  journal={arXiv preprint arXiv:2411.12783},
  year={2024}
}

@inproceedings{banerjee2024rexamine,
  title={ReXamine-Global: A framework for uncovering inconsistencies in radiology report generation metrics},
  author={Banerjee, Oishi and Saenz, Agustina and Wu, Kay and Clements, Warren and Zia, Adil and Buensalido, Dominic and Kavnoudias, Helen and Abi-Ghanem, Alain S and Ghawi, Nour El and Luna, Cibele and others},
  booktitle={Biocomputing 2025: Proceedings of the Pacific Symposium},
  pages={185--198},
  year={2024},
  organization={World Scientific}
}

@article{wang2023r2gengpt,
  title={R2gengpt: Radiology report generation with frozen llms},
  author={Wang, Zhanyu and Liu, Lingqiao and Wang, Lei and Zhou, Luping},
  journal={Meta-Radiology},
  volume={1},
  number={3},
  pages={100033},
  year={2023},
  publisher={Elsevier}
}

@inproceedings{chen2025dia,
  title={Dia-LLaMA: Towards large language model-driven ct report generation},
  author={Chen, Zhixuan and Luo, Luyang and Bie, Yequan and Chen, Hao},
  booktitle={International Conference on Medical Image Computing and Computer-Assisted Intervention},
  pages={141--151},
  year={2025},
  organization={Springer}
}

@inproceedings{irvin2019chexpert,
  title={Chexpert: A large chest radiograph dataset with uncertainty labels and expert comparison},
  author={Irvin, Jeremy and Rajpurkar, Pranav and Ko, Michael and Yu, Yifan and Ciurea-Ilcus, Silviana and Chute, Chris and Marklund, Henrik and Haghgoo, Behzad and Ball, Robyn and Shpanskaya, Katie and others},
  booktitle={Proceedings of the AAAI conference on artificial intelligence},
  volume={33},
  number={01},
  pages={590--597},
  year={2019}
}

@misc{zhao2024ratescore,
  title={Ratescore: A metric for radiology report generation. medRxiv},
  author={Zhao, Weike and Wu, Chaoyi and Zhang, Xiaoman and Zhang, Ya and Wang, Yanfeng and Xie, Weidi},
  year={2024},
  publisher={2024a}
}

@article{zhang2019bertscore,
  title={Bertscore: Evaluating text generation with bert},
  author={Zhang, Tianyi and Kishore, Varsha and Wu, Felix and Weinberger, Kilian Q and Artzi, Yoav},
  journal={arXiv preprint arXiv:1904.09675},
  year={2019}
}

@article{jain2021radgraph,
  title={Radgraph: Extracting clinical entities and relations from radiology reports},
  author={Jain, Saahil and Agrawal, Ashwin and Saporta, Adriel and Truong, Steven QH and Duong, Du Nguyen and Bui, Tan and Chambon, Pierre and Zhang, Yuhao and Lungren, Matthew P and Ng, Andrew Y and others},
  journal={arXiv preprint arXiv:2106.14463},
  year={2021}
}

@inproceedings{ostmeier2024green,
  title={Green: Generative radiology report evaluation and error notation},
  author={Ostmeier, Sophie and Xu, Justin and Chen, Zhihong and Varma, Maya and Blankemeier, Louis and Bluethgen, Christian and Md, Arne Edward Michalson and Moseley, Michael and Langlotz, Curtis and Chaudhari, Akshay S and others},
  booktitle={Findings of the association for computational linguistics: EMNLP 2024},
  pages={374--390},
  year={2024}
}

@inproceedings{hamamci2025crg,
  title={CRG Score: A Distribution-Aware Clinical Metric for Radiology Report Generation},
  author={Hamamci, Ibrahim Ethem and Er, Sezgin and Shit, Suprosanna and Reynaud, Hadrien and Kainz, Bernhard and Menze, Bjoern},
  booktitle={Medical Imaging with Deep Learning-Short Papers},
  year={2025}
}

@article{hamamci2024foundation,
  title={A foundation model utilizing chest ct volumes and radiology reports for supervised-level zero-shot detection of abnormalities},
  author={Hamamci, Ibrahim Ethem and Er, Sezgin and Almas, Furkan and Simsek, Ayse Gulnihan and Esirgun, Sevval Nil and Dogan, Irem and Dasdelen, Muhammed Furkan and Wittmann, Bastian and Simsar, Enis and Simsar, Mehmet and others},
  journal={CoRR},
  year={2024}
}

@article{yu2023evaluating,
  title={Evaluating progress in automatic chest x-ray radiology report generation},
  author={Yu, Feiyang and Endo, Mark and Krishnan, Rayan and Pan, Ian and Tsai, Andy and Reis, Eduardo Pontes and Fonseca, Eduardo Kaiser Ururahy Nunes and Lee, Henrique Min Ho and Abad, Zahra Shakeri Hossein and Ng, Andrew Y and others},
  journal={Patterns},
  volume={4},
  number={9},
  year={2023},
  publisher={Elsevier}
}

\end{document}